\documentclass{article}
\usepackage[preprint, nonatbib]{nips_2018}
\usepackage{float}
\usepackage[square,numbers,sort&compress]{natbib}
\usepackage[utf8]{inputenc}
\usepackage[T1]{fontenc}
\usepackage{hyperref}
\usepackage{url}
\usepackage{booktabs}
\usepackage{amsfonts}
\usepackage{nicefrac}
\usepackage{microtype}
\usepackage{graphicx}
\usepackage{tabularx}
\usepackage{amsmath}
\usepackage{array}
\newcolumntype{Y}{>{\raggedright\arraybackslash}X}

\newcolumntype{L}{>{\raggedright\arraybackslash}X}

\DeclareUnicodeCharacter{2460}{\textcircled{\scriptsize{1}}}
\DeclareUnicodeCharacter{2461}{\textcircled{\scriptsize{2}}}
\DeclareUnicodeCharacter{2462}{\textcircled{\scriptsize{3}}}
\DeclareUnicodeCharacter{2463}{\textcircled{\scriptsize{4}}}
\DeclareUnicodeCharacter{2464}{\textcircled{\scriptsize{5}}}
\DeclareUnicodeCharacter{2465}{\textcircled{\scriptsize{6}}}
\DeclareUnicodeCharacter{2466}{\textcircled{\scriptsize{7}}}
\DeclareUnicodeCharacter{2467}{\textcircled{\scriptsize{8}}}

\title{Flexible Parallel Neural Network Architecture Model for Early Prediction of Lithium Battery Life}
\author{
  Lidang Jiang, Zhuoxiang Li, Changyan Hu, Qingsong Huang*, Ge He* \\
  School of Chemical Engineering, Sichuan University, Chengdu 610000 Sichuan, PR China \\
  \texttt{hege@scu.edu.cn, qshuang@scu.edu.cn}
}
\usepackage{graphicx}

\newcommand{\includegraphicsfit}[2][]{
  \includegraphics[width=\textwidth, height=\textheight, keepaspectratio, #1]{#2}
}

\usepackage{indentfirst} 

\usepackage{titlesec}

\titlespacing*{\section}
  {0pt} 
  {2\parskip} 
  {2\parskip} 

\titlespacing*{\subsection}
  {0pt} 
  {2\parskip} 
  {2\parskip} 

\titlespacing*{\subsubsection}
  {0pt} 
  {2\parskip} 
  {2\parskip} 

\begin{document}

\maketitle
\begin{abstract}
The early prediction of battery life (EPBL) is vital for enhancing the efficiency and extending the lifespan of lithium batteries. Traditional models with fixed architectures often encounter underfitting or overfitting issues due to the diverse data distributions in different EPBL tasks. An interpretable deep learning model of flexible parallel neural network (FPNN) is proposed, which includes an InceptionBlock, a 3D convolutional neural network (CNN), a 2D CNN, and a dual-stream network. The proposed model effectively extracts electrochemical features from video-like formatted data using the 3D CNN and achieves advanced multi-scale feature abstraction through the InceptionBlock. The FPNN can adaptively adjust the number of InceptionBlocks to flexibly handle tasks of varying complexity in EPBL. The test on the MIT dataset shows that the FPNN model achieves outstanding predictive accuracy in EPBL tasks, with MAPEs of 2.47\%, 1.29\%, 1.08\%, and 0.88\% when the input cyclic data volumes are 10, 20, 30, and 40, respectively. The interpretability of the FPNN is mainly reflected in its flexible unit structure and parameter selection: its diverse branching structure enables the model to capture features at different scales, thus allowing the machine to learn informative features. The approach presented herein provides an accurate, adaptable, and comprehensible solution for early life prediction of lithium batteries, opening new possibilities in the field of battery health monitoring.
\end{abstract}

\setlength{\parindent}{2em}

\section{ Introduction}
Since Sony introduced the first commercial Lithium-Ion Batteries (LIBs) in 1991\cite{yoshino2012birth}, these batteries have been widely used in portable electronic devices and electric vehicles\cite{gan2019enhancing, teki2009nanostructured} due to their long service life and low self-discharge rate\cite{sehrawat2021recent}. Early Prediction of Battery Life (EPBL) is not only crucial for guiding improvements in product design, enhancing safety, optimizing maintenance strategies, and conducting cost-benefit analysis, but is also significant to battery manufacturers, engineers, and end-users.

There are primarily two approaches to modeling EPBL: physics-based modeling and data-driven modeling. Physics-based methods include semi-empirical models\cite{han2014comparative}, empirical models\cite{lin2020battery,ma2021remaining}, equivalent circuit models\cite{fleischer2014line,ouyang2020online}, and electrochemical models\cite{allam2020online,li2020electrochemical,liu2019degradation}. These methods require a deep understanding of the physical and chemical processes in the domain of batteries, and use mathematical models to predict battery life. Physics-based modeling excels in explaining battery behavior and considering the interactions of various influencing factors, thus offering advantages in model accuracy and interpretability. However, this approach often requires extensive experimental data for parameter fitting and calibration, making the modeling process time-consuming and the model structure complex. Additionally, the theoretical assumptions of physics-based models may not be entirely applicable in complex real-world scenarios\cite{severson2019data}.

Compared to physics-based modeling, data-driven modeling\cite{wang2016remaining}, particularly machine learning (ML)\cite{jordan2015machine}, demonstrates superior performance in terms of flexibility, adaptability, and scalability. Various ML methods have been applied to the battery life prediction (BLP), such as using support vector machine (SVM) to identify batteries with abnormal life spans\cite{saxena2018anomaly}, and employing decision trees\cite{myles2004introduction}, SVM \cite{noble2006support}, and k-nearest neighbors (KNN) \cite{guo2003knn} for classifying the life of Lithium Iron Phosphate (LFP) /graphite batteries\cite{zhu2019predicting}. Severson et al. \cite{severson2019data} trained a simple linear model on the MIT (Massachusetts Institute of Technology) battery dataset, achieving a remaining useful life (RUL) prediction accuracy of 9.1\%. Zhang et al. \cite{zhang2020identifying} constructed an electrochemical impedance spectroscopy (EIS) dataset and achieved high-precision RUL prediction using gaussian process regression (GPR). Yang et al. \cite{yang2020lifespan} used a gradient boosting regression tree model based on features such as voltage, capacity, and temperature, achieving a mean absolute percentage error (MAPE) of 7\% in EPBL tasks. Fei et al. \cite{fei2021early} developed a comprehensive ML framework that includes feature extraction, feature selection, and a prediction module, which integrates various ML models such as GPR, SVM, and Elastic Net (linear regression). Their results showed that SVM achieved the lowest MAPE of 8.0\% on the first 100 cycles of the MIT dataset. However, these traditional ML algorithms typically require manual selection and design of features, which adds complexity to the ML workflow.

With the advancement in computational power, deep learning methods, epitomized by neural networks\cite{lecun2015deep}, have shown tremendous potential in the field of battery life prediction \cite{zhang2021prognostics}. The capacity of lithium-ion batteries gradually diminishes over time, leading to reduced lifespan. Recurrent Neural Networks (RNN) \cite{zhao2020lithium} and their variants\cite{hong2020online, ren2020data, wei2020state,yi2023digital}, known for their excellence in handling time-series data and capturing long-term dependencies within the data, have been widely applied. However, these methods overlook the spatial characteristics of battery data. By preprocessing lithium battery feature data into an image-like format, 2D Convolutional Neural Networks (CNN) can effectively extract crucial spatial features, thereby enhancing the accuracy of battery state prediction \cite{bian2022robust,chen2022novel,gong2021method, li2021lithium}. On the other hand, transforming battery data into a video-like format and processing it with 3D CNN allows for the comprehensive capture of complex relationships between voltage, current, and temperature, facilitating the integration of different electrochemical characteristics. Consequently, Yang \cite{yang2021machine} proposed a hybrid CNN model that significantly improved prediction accuracy, reducing the prediction error to 3.08\% with only 20 charging cycles of data.

However, different EPBL tasks have varying data distributions, demanding models of different complexities to avoid underfitting or overfitting. Herein, we designed the Flexible Parallel Neural Network (FPNN), a neural network model capable of rapid and adaptive adjustment. The core of FPNN is a combination of Inception-ResNet-A\cite{szegedy2017inception}, 3D CNN \cite{ji20123d}, 2D CNN, and a dual-stream network \cite{simonyan2014two}. After preprocessing, each sample forms a video-like format and then the 3D CNN integrates temporal (depth, charging capacity index) and spatial (channel, voltage/current/temperature) features to obtain the primary features. These primary features are extracted at a higher level through the InceptionBlock (i.e., Inception-ResNet-A module), forming multi-scale features. Multiple InceptionBlock flexible units are stacked to form an InceptionBlocks structure., and residual connections are used to further enhance the model's ability to extract electrochemical features. Additionally, the number of InceptionBlocks is learned automatically by the model, enabling FPNN to flexibly adapt to different EPBL tasks.

The main contributions of this paper are as follows:

1) Adaptability and Accuracy: The novel FPNN model can rapidly and flexibly adapt various EPBL tasks, achieving high prediction accuracy. FPNN demonstrates excellent performance with both fixed and flexible architectures.

2) Interpretability: This is primarily reflected in its flexible unit structure and parameter selection. By implementing rapid integration of different channel information through 1×1 convolution kernels, FPNN effectively consolidates diverse information. Additionally, its varied branch structure enables the model to capture features at different scales, allowing the machine to learn a rich set of feature information. This structural design not only enhances the model's predictive performance but also makes its decision-making process more transparent and understandable.

The article is organized as follows: Section 2 introduces the MIT dataset, highlighting its relevance to EPBL tasks. Section 3 outlines the FPNN workflow, focusing on data preprocessing and architectural design. Section 4 discusses experimental results, comparing FPNN's performance in EPBL tasks with other existing models, and explores the model's interpretability, concluding with key findings. Finally, conclusions are drawn.

\section{ Dataset}
The dataset for modeling herein is the MIT dataset created by Severson et al. \cite{severson2019data} in 2019, aimed at studying the rapid charging issues of batteries. This dataset includes 124 A123 lithium iron phosphate/graphite batteries, tested in a constant temperature environment of 30°C. However, due to internal electrochemical reactions in the batteries, the actual temperature fluctuates around 30°C. As shown in Figure 1a, the nominal capacity of the batteries in the dataset is 1.1 Ah. A battery is defined as failed when its discharge capacity falls to 80\% of the nominal capacity, and the corresponding cycle number is defined as the battery life. The dataset primarily follows the "C1(Q1)-C2" charging strategy, which involves initial constant current charging with C1 current until the charge reaches Q1 (State of Charge, SOC, \%), then switching to C2 current for continued constant current charging until SOC reaches 80\%, and finally charging at 1C CC-CV (Constant Current-Constant Voltage), where 1C refers to a charging rate that would fully charge the battery in one hour, until the cut-off voltage of 3.6 V.with a voltage lower limit of 2.0 V. By varying the values of C1, C2, and Q1, 72 different charging strategies were generated to simulate a variety of charging modes in real-world environments.

In practical applications, the way batteries are used (i.e., discharged) varies greatly depending on the device and usage conditions, making the discharge process complex and varied. In contrast, the way batteries are charged (i.e., the charging process) is more standardized and consistent.Therefore, this study chooses to use only charging data as the model input to avoid randomness, which better adapts to real-world application scenarios. In the MIT dataset, there is a mapping relationship between the variation of charging current and battery life: as the number of cycles increases, the battery voltage shows a downward trend in the early stages of charging and an upward trend in the later stages, as shown in Figure 1b. This reflects the decline in battery capacity, which has a significant impact on battery life. Additionally, Figure 1c shows the variation of temperature with the number of cycles, revealing the changes in the internal electrochemical characteristics of the battery. Therefore, the voltage, current, and temperature data from different cycles of different batteries can serve as samples containing specific electrochemical information.

\begin{figure}[H]
  \centering
  \includegraphicsfit{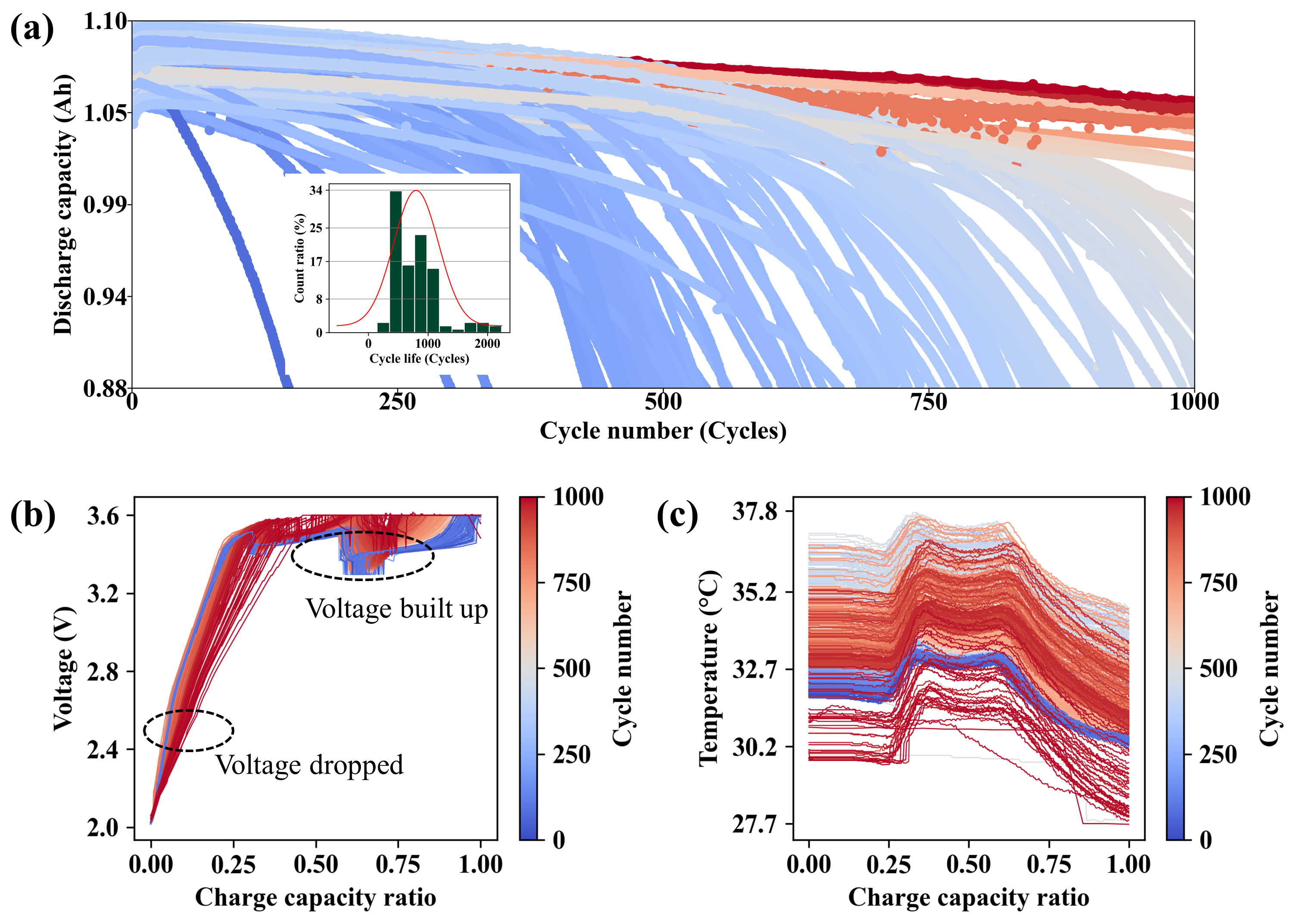}
  \caption{provides a comprehensive analysis of lithium-ion battery performance: (a) Based on the MIT dataset, it shows the trend of lithium-ion battery discharge capacity decay over cycles, with an inset graph revealing the statistical distribution of battery life. (b) Displays the voltage changes of the "b1c23" battery across various charging cycles. The areas of voltage rise and fall are marked with black circles in the graph, highlighting the fluctuation characteristics of voltage during the charging process. (c) Describes the temperature variation trend of the "b1c23" battery during the charging process, where the temperature changes reflect the thermal management state at different charging stages.}
  \label{fig:image1}
\end{figure}

\section{ Methods}
Following the detailed introduction of the dataset, this section will present the complete workflow for EPBL tasks. As shown in Figure 2, the process begins with the Battery Management System (BMS), which is responsible for collecting operational data of the battery. Subsequently, these raw data undergo a series of preprocessing steps and are transformed into a video-like format to enhance their processability. The transformed data are then fed into the FPNN for training the model and performing prediction tasks. The hyperparameters of FPNN are determined through Bayesian optimization algorithm, and except for the Number of InceptionBlocks (NOI), the hyperparameters for different EPBL tasks are the same. The plot that compares the actual and the predicted data provides users with an intuitive performance assessment.

\begin{figure}[H]
  \centering
  \includegraphicsfit{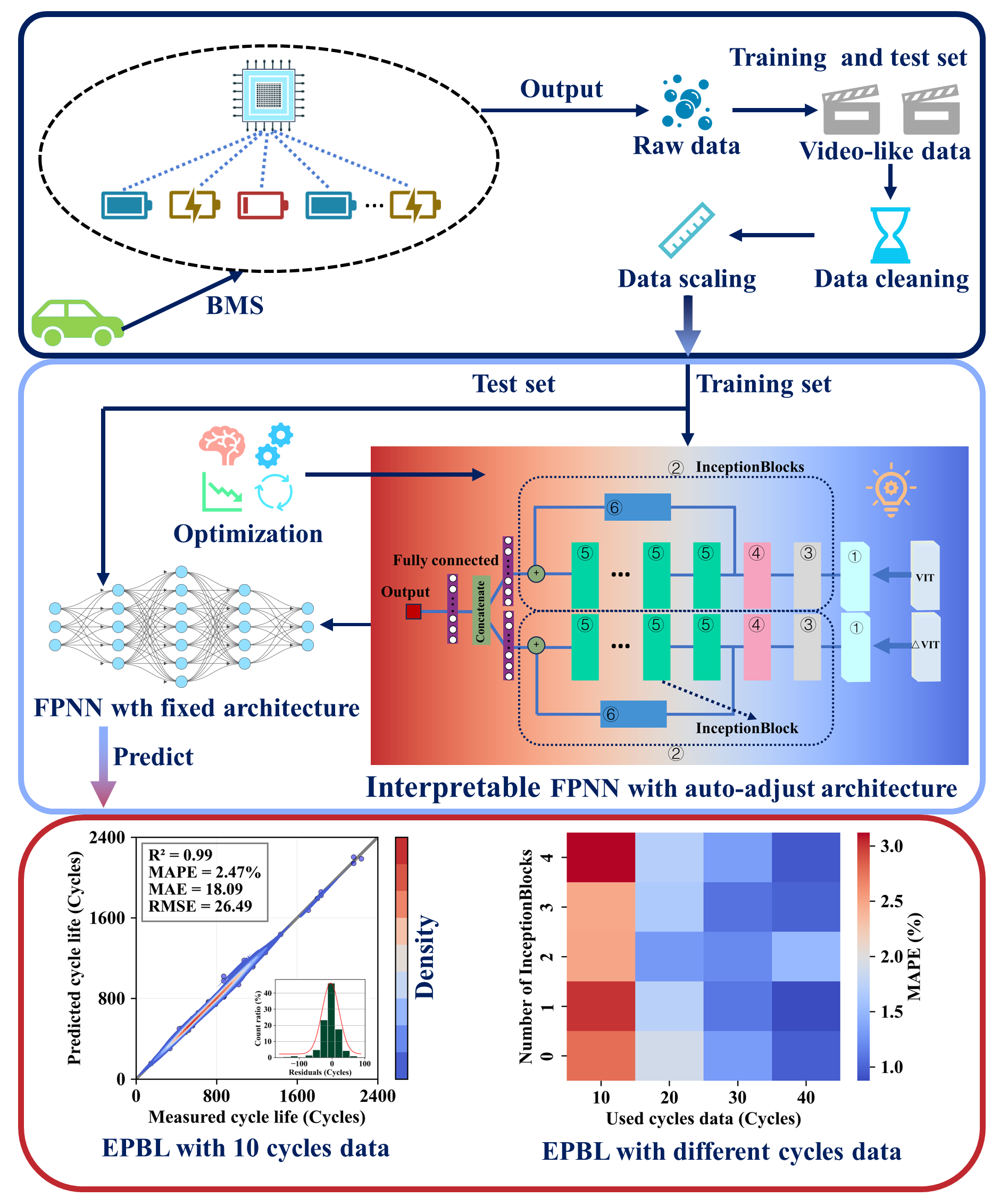}
  \caption{Schematic Diagram of the EPBL Technology Route Based on FPNN}
  \label{fig:image2}
\end{figure}

\subsection{ Data Preprocessing}
Before being input into the model, the data undergo a series of preprocessing steps. The selected data include voltage, current, and temperature from the charging part of the dataset, which are organized into three matrix blocks. Taking voltage as an example, the horizontal axis of the data indicates the number of cycles, while the vertical axis refers to the charging capacity. As shown in Figure 3a, each sample consists of the first cycle of the battery and its three most recent cycles. This method of sample selection, different from other studies\cite{yang2021machine,zhang2022deep}, is a result of balancing the number of samples with feature similarity. After the samples are organized into a video-like format, they are divided into training and test sets in a 94:30 ratio, with outliers removed and smoothed using a Savitzky-Golay filter. The features of the training and test sets are scaled to a range of -1 to 1, while the labels remain unchanged, and then are directly input into one branch of the FPNN. Additionally, considering that differential data are also important features, the voltage, current, and temperature data of the battery are differenced with their corresponding first cycle data to obtain differential features. The processing of these differential features is similar to that of normal features, and they are finally input into another branch of the FPNN.

\subsection{ FPNN (Flexible Parallel Neural Network)}
In this study, the Bayesian optimization algorithm \cite{snoek2012practical} was employed to determine the hyperparameters of the developed FPNN. Bayesian optimization uses Gaussian process regression as a surrogate model, which provides both the predicted value and its confidence interval at a given point, aiding in more effectively balancing exploration and exploitation. The NOI is a key hyperparameter that endows FPNN with the adaptability to different EPBL tasks. We searched for the optimal hyperparameters combination under the condition of 10 cycles of data. To accurately assess the impact of NOI on the predictive performance of FPNN, when exploring the performance with other cycles of data, all hyperparameters except NOI were kept the same as in the case of 10 cycles. This approach ensures the validity and comparability of the experimental results.
\subsubsection{ FPNN Architecture}
The FPNN primarily consists of a dual-stream network, a 3D CNN, and a flexible unit InceptionBlock. The dual-stream network processes different types of features. The 3D CNN is responsible for integrating voltage, current, and temperature (VIT) features and passing low-level features to the next layer. Inception-ResNet-A, after simplification and modification, forms the InceptionBlock: it reduces the number of channels in feature maps to lower memory usage, thereby decreasing the hardware requirements for model deployment. Multiple InceptionBlock flexible units, along with the initial layers, constitute the InceptionBlocks. To ensure that the features extracted by the initial layers are effectively transmitted between different InceptionBlocks, residual connections have been added in this study.

\begin{figure}[H]
  \centering
  \includegraphicsfit{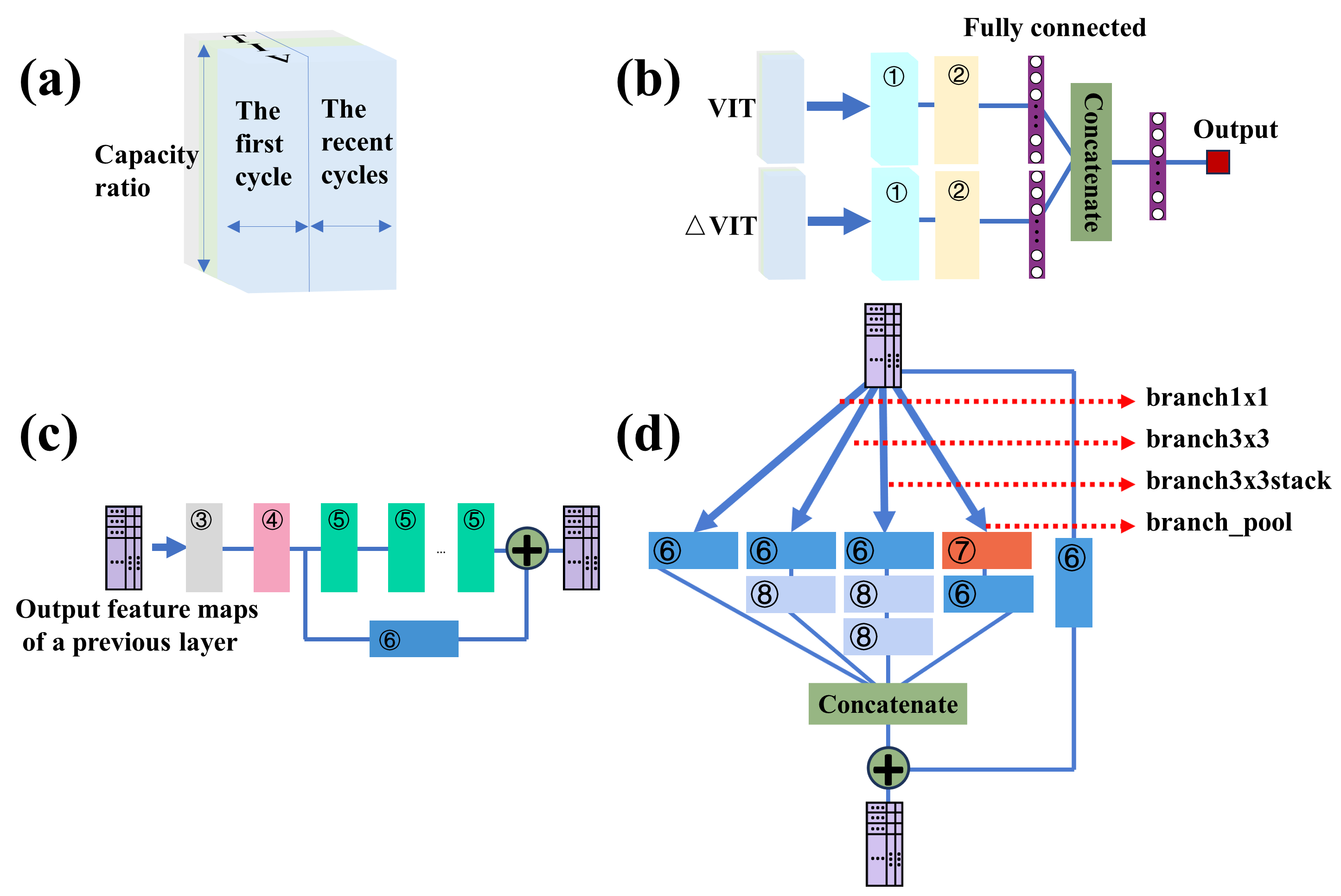}
  \caption{describes the detailed architecture and components of FPNN: ① 3D convolutional layer, using 3×3 convolutional kernels, with 64 channels; ② InceptionBlocks module; ③ 2D convolutional layer, with a kernel size of 7×7 and 64 channels; ④ Max pooling layer, with a pooling kernel size of 3×3; ⑤ InceptionBlock flexible unit; ⑥ 2D convolutional layer, with a kernel size of 1×1 and 16 or 24 channels (used as the target channel number for residual connections in other cases); ⑦ Average pooling layer, with a pooling kernel size of 3×3; ⑧ 2D convolutional layer, with a kernel size of 3×3 and 16 or 24 channels. The figure also shows: (a) FPNN video-like data after preprocessing; (b) The overall architecture of FPNN; (c) Detailed structure of the flexible module InceptionBlocks; (d) Specific details of the flexible unit InceptionBlock.}
  \label{fig:image3}
\end{figure}

\subsubsection{ FPNN Units in FPNN}
The detailed structure of each flexible unit, the InceptionBlock, is shown in Figure 3d. The InceptionBlock takes the output of the previous layer as input and is then divided into four branches, each consisting of 2D CNNs with different hyperparameters. The operations of the 2D CNN include convolution (1), normalization, and the activation calculation of Leaky ReLU (2) for non-linearization, thereby extracting the non-linear electrochemical information.
\begin{equation}
C(i, j)=\sum_m \sum_n I(i+m, j+n) \cdot K(m, n)
\end{equation}   
where $I(i+m, j+n)$ is the element in the input matrix; $K(m, n)$ is the element in the convolution kernel; $m$ and $n$ are the row and column indices of the convolution kernel; and $C(i, j)$ is the element in the convolution output.
\begin{equation}
f(x)= \begin{cases}x & \text { if } x>0 \\ \alpha x & \text { if } x \leq 0\end{cases}
\end{equation}
where $x$ represents the input to the activation function; $f(x)$ denotes the output of the activation function; $\alpha$ is a small constant, typically ranging between 0 and 1.

The first branch contains a 2D convolutional layer with a 1×1 convolution kernel and 16 channels; the second branch is composed of a 2D convolutional layer with a 1×1 convolution kernel and 16 channels, followed by a 2D convolutional layer with a 3×3 convolution kernel and 24 channels; the third branch consists of a 2D convolutional layer with a 1×1 convolution kernel and 16 channels, and two 2D convolutional layers with a 3×3 convolution kernel and 24 channels; the fourth branch is a sequential model made up of an average pooling layer with a 3×3 pooling kernel (3) and a 2D convolutional layer with a 3×3 convolution kernel and 24 channels.
\begin{equation}
P(i, j)=\frac{1}{K \times L} \sum_{m=0}^{K-1} \sum_{n=0}^{L-1} I(i \times S+m, j \times S+n)
\end{equation}
where $P(i, j)$ represents the elements in the pooling output; $I(i \times S+m, j \times S+n)$ denotes the elements in the input feature map; $(K, L)$ is the size of the pooling window; $K$ is the height of the window; and $L$ is the width of the window. $S$ is the stride, controlling the step length of the window's movement. $(i, j)$ is the index in the output feature map. $(m, n)$ is the relative index within the pooling window. The outputs of these four branches are concatenated together, forming the output features of the InceptionBlock. To ensure effective transmission of the features extracted by the previous layer across each layer of the InceptionBlock, residual connections (4) are used, which include a linear transformation to adjust the dimensions of the input and output for matching.
\begin{equation}
R(x)=F(x)+W x
\end{equation}
where $F(x)$ is the output of a series of operations (such as convolution, activation, etc.) in the residual block; and $R(x)$ is the output of the residual block; $W x$ is the input to the residual block. Since the dimensions of $F(x)$ and $x$ do not match, a linear transformation $W$ is added here to adjust the dimensions of $x$, making it match with $F(x)$.

As shown in Figure 3c, the flexible module InceptionBlocks consists of multiple InceptionBlock flexible units. The output of the previous layer first undergoes preliminary feature extraction through a 2D convolutional layer with a 7×7 convolution kernel and 64 channels, and then the output of this layer serves as the input for the max pooling layer with a 3×3 pooling kernel (5).
\begin{equation}
P(i, j)=\max _{0 \leq m<K, 0 \leq n<L} I(i \times S+m, j \times S+n)
\end{equation}
where $P(i, j)$ represents the elements in the pooling output, $I(i \times S+m, j \times S+n)$ denotes the elements in the input feature map, $(K, L)$ is the size of the pooling window, $K$ is the height of the window, and $L$ is the width of the window. $S$ is the stride, controlling the step length of the window's movement. $(i, j)$ is the index in the output feature map. $(m, n)$ is the relative index within the pooling window. This combination of the $2 \mathrm{D}$ convolutional layer and the max pooling layer is defined as the initial layer of InceptionBlocks, aimed at rapid feature extraction and the feature dimensionality reduction, preparing for subsequent feature extraction in InceptionBlock. The output of the max pooling layer serves as the input to the first InceptionBlock, and then the output of the first InceptionBlock is passed to the next layer. The NOI in InceptionBlocks is not fixed and may require different levels of network complexity for different EPBL.

To comprehensively evaluate the predictive performance of the model, this study selects the MAPE, Mean Absolute Error (MAE), and Root Mean Square Error (RMSE) as evaluation metrics. The corresponding mathematical expressions are given in equations (6) to (8).
\begin{equation}
\begin{aligned}
\mathrm{MAPE} & =\frac{100 \%}{n} \sum_{i=1}^n\left|\frac{y_i-\hat{y}_i}{y_i}\right|
\end{aligned}
\end{equation}
\begin{equation}
\begin{aligned}
\mathrm{MAE} & =\frac{1}{n} \sum_{i=1}^n\left|y_i-\hat{y}_i\right|
\end{aligned}
\end{equation}
\begin{equation}
\begin{aligned}
\mathrm{RMSE} & =\sqrt{\frac{1}{n} \sum_{i=1}^n\left(y_i-\hat{y}_i\right)^2}
\end{aligned}
\end{equation}
where $n$ is the total number of samples, $y_i$ is the actual value of the $i$ sample, and $\hat{y}_i$ is the predicted value of the $i$ sample.
\section{ Results and Discussion}
\subsection{ Evaluation of FPNN Model Predictive Performance }
This section evaluates the predictive performance of the FPNN on different cyclic data for EPBL, with a particular focus on the impact of NOI. Initially, the study found that different settings of NOI significantly affect the prediction accuracy of FPNN in EPBL tasks. As shown in Table 1, taking 10 cycles of data as an example, when NOI increases from 1 to 3, the MAPE gradually decreases to 2.47\%, indicating that appropriately increasing NOI can enhance predictive performance. However, when NOI increases to 4, the MAPE rises again to 3.12\%, possibly due to the excessive depth of the network leading to a reduced capacity for transmitting nonlinear information.

Furthermore, experimental results also indicate that as the volume of training data increases, the optimal value of NOI gradually decreases. For instance, when using 10, 20, 30, and 40 cycles of data, the optimal NOI values are 3, 2, 1 (3), and 1, respectively, with MAPEs of 2.47\%, 1.29\%, 1.10\% (1.08\%), and 0.88\%, respectively. This suggests that in cases of large data volumes, a more simplified FPNN structure is sufficient to achieve high prediction accuracy. This trend implies that as the volume of data increases, some complex parts of the FPNN may become redundant, or even negatively impact the model's predictive performance.

\begin{table}[H]
  \centering
  \caption{The Impact of NOI on FPNN's EPBL}
  \begin{tabularx}{\textwidth}{X X X X X}
    \toprule
    Dataset & Blocks & MAPE (\%) & MAE (cycles) & RMSE (cycles) \\
    \midrule
    10 Cycles  & 0 & 2.77 & 21.56 & 31.65 \\
    & 1 & 3.01 & 22.84 & 33.88 \\ 
    & 2 & 2.50 & 18.98 & 27.16 \\ 
    & 3 & 2.47 & 18.09 & 26.49 \\ 
    & 4 & 3.12 & 23.95 & 35.68 \\  
    \midrule
    20 Cycles  & 0 & 1.89 & 14.81 & 21.55 \\  
    & 1 & 1.72 & 13.28 & 18.30 \\ 
    & 2 & 1.29 & 10.60 & 17.60 \\ 
    & 3 & 1.66 & 12.76 & 19.98 \\ 
    & 4 & 1.71 & 13.18 & 22.36 \\  
    \midrule
    30 Cycles  & 0 & 1.31 & 10.25 & 15.02 \\ 
    & 1 & 1.10 & 8.52 & 12.95 \\ 
    & 2 & 1.21 & 8.59 & 12.85 \\ 
    & 3 & 1.08 & 8.04 & 12.48 \\ 
    & 4 & 1.33 & 10.73 & 19.68 \\ 
    \midrule
    40 Cycles  & 0 & 0.97 & 7.59 & 11.45 \\
    & 1 & 0.88 & 6.81 & 10.03 \\ 
    & 2 & 1.51 & 12.26 & 17.05\\ 
    & 3 & 1.00 & 7.87 & 11.90 \\ 
    & 4 & 0.94 & 7.26 & 10.90 \\  
    \bottomrule
  \end{tabularx}
\end{table}

Figure 4a shows the performance of the FPNN trained with 10 cycles of data in the EPBL task, with an overall MAPE of 2.47\%. Although individual samples exhibit larger prediction errors, the majority have small errors, demonstrating high overall prediction accuracy. This confirms the effectiveness of the FPNN model, especially under conditions of limited data.

From Figure 4b, it is evident how different volumes of cyclic data and NOI settings affect the FPNN's prediction MAPE. The results show that as the volume of cyclic data increases, the EPBL prediction accuracy of the FPNN significantly improves. Particularly when NOI is fixed at 3, the model's MAPE remains at a low level regardless of the amount of cyclic data, proving the adaptability and stability of the model structure. When the input cyclic data volumes are 10, 20, 30, and 40, respectively, even if NOI is not at its optimal value, the model's MAPEs are still 2.47\%, 1.66\%, 1.08\%, and 1.00\%, respectively. Under the optimal NOI settings, the corresponding MAPEs are further reduced to 2.47\%, 1.29\%, 1.08\%, and 0.88\%.

Figure 4c displays the trend of EPBL prediction accuracy with changes in the volume of cyclic data and NOI. Under the optimal NOI settings, the fluctuation in test errors is small, and the median error is low. As the volume of cyclic data increases, the range of error fluctuation noticeably decreases, and the median error also reduces, further emphasizing the impact of data volume on prediction accuracy.

Figure 4d-e further confirm the robustness and stability of the FPNN model. Across different lifespan intervals of sample distribution, the model can handle samples of varying frequencies, demonstrating outstanding performance. Particularly in the tests on two samples with different frequencies, 'b1c3' and 'b2c20', the FPNN model shows that while there is randomness in the prediction error for individual samples, the overall performance remains superior.

Finally, according to Table 2, the method proposed in this study shows significant advantages compared to other published EPBL prediction methods. The FPNN models using 10 and 40 cycles of data have MAPEs of 2.47\% and 0.88\%, respectively, which are significantly lower than the MAPEs of methods based on SVM, Linear model, and GBRT. Additionally, the number of early cycles required is greatly reduced, highlighting the efficiency of FPNN in terms of data utilization.

\begin{figure}[H]
  \centering
  \includegraphicsfit{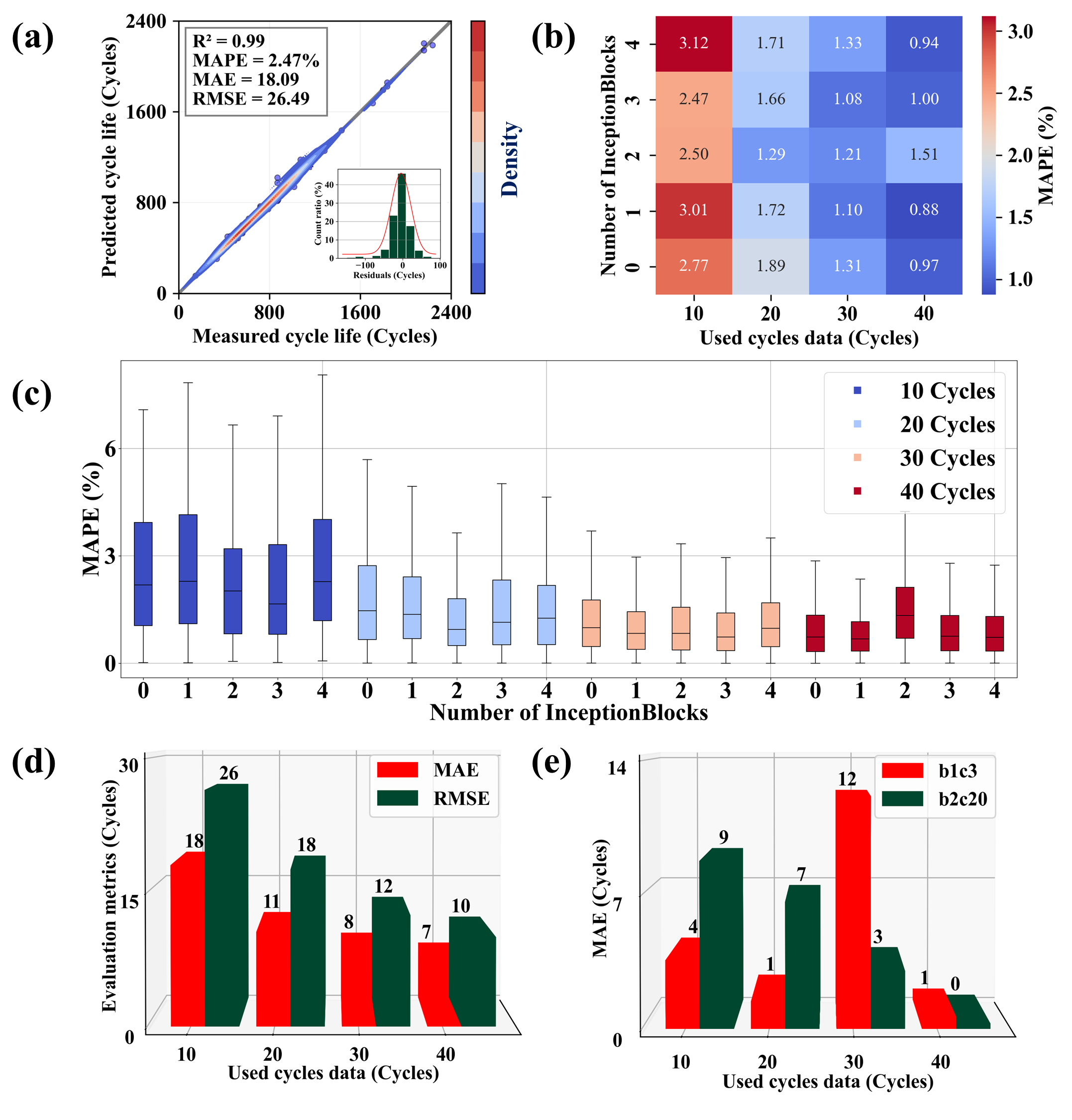}
  \caption{comprehensively demonstrates the performance of FPNN in EPBL tasks: (a) shows the EPBL results of the FPNN trained with the first 10 cycles of data, incorporating a subplot that displays a histogram of residual frequencies; (b) is a heatmap illustrating the impact of different cyclic data volumes and NOI settings on the FPNN's prediction MAPE; (c) a box plot, further revealing the distribution of prediction errors; (d) shows the MAE and RMSE of FPNN predictions at different cyclic data volumes; (e) displays the MAE performance for specific batteries 'b1c3' and 'b2c20'.}
  \label{fig:image4}
\end{figure}
\indent 

\begin{table}[H]
  \centering
  \caption{EPBL of Other Published Methods}
  \begin{tabularx}{\textwidth}{YYYY}
    \toprule
    Methods & Used Early Cycles & MAPE(\%) & RMSE(Cycles) \\
    \midrule
    SVM\cite{fei2021early} & 100 & 8.00 & 115.00 \\
    Linear \cite{severson2019data} & 100 & 7.50 & 100.00 \\
    GBRT\cite{yang2020lifespan} & 250 & 7.00 & 82.80 \\
    HCNN\cite{yang2021machine} & 20 & 3.08 & 42.00 \\
     & 60 & 1.12 & 13.00 \\
    \textbf{FPNN} & \textbf{10} & \textbf{2.47} & \textbf{26.49} \\
     & \textbf{40} & \textbf{0.88} & \textbf{10.03} \\
    \bottomrule
  \end{tabularx}
\end{table}

Overall, when the data volume is small, the FPNN requires a more complex structure to effectively extract features, hence its optimal NOI value is relatively high. As the data volume increases, the FPNN model can improve its predictive performance by simplifying its structure. This indicates that on large datasets, some complex network structures may no longer be necessary, and a simplified FPNN is sufficient to achieve good predictive results. This also suggests that in the context of big data, complex structures may negatively impact model performance.

\subsection{ Ablation Experiments}
To delve deeper into the role of each module of the FPNN and reveal their contributions to prediction accuracy, this paper conducted ablation experiments, with results shown in Table 3 and Figure 5. Table 3 summarizes the specific data regarding the impact of removing certain components of FPNN under the optimal NOI settings on performance. Figure 5 visually presents the data from the ablation experiments, where NaN values are replaced with 0 in the graph. Due to the large differences in extreme values of the data, a simple mathematical transformation was applied to the original data, namely $y=\log (1+x)$, where  represents the error evaluation metrics, and  is the value on the vertical axis in the graph.

Under the condition of using only 10 cycles of data, the results show that removing the differential feature branch leads to a surge in MAPE to 98.92\%, and the MAE and RMSE also increase to 772.33 cycles and 834.04 cycles, respectively. This highlights the critical role of the differential feature branch in enhancing model accuracy. When the residual connections are removed, the MAPE slightly rises to 2.58\%, and the MAE and RMSE also increase accordingly, revealing the importance of residual connections in reducing prediction errors. After removing the 3D convolutional layer, the MAPE rises to 3.09\%, further confirming the significance of the 3D convolutional layer in feature extraction. Finally, when the initial layers is removed, the MAPE increases to 3.45\%, reflecting the important role of the initial layers in model stability.

When using 20, 30, and 40 cycles of data, the results of the ablation experiments reveal similar patterns. Particularly in the absence of the initial layers, the model training could not proceed normally, resulting in NaN values. Notably, with 40 cycles of data, since the optimal NOI value is 1, removing the residual connections leads to a slight decrease in MAPE, indicating that under certain conditions, the 1×1 2D convolution used for rapid channel number conversion had side effects, and the residual connections might not be necessary. The redundancy in the FPNN structure led to a decrease in the accuracy of the EPBL task, reaffirming the conclusions drawn in section 4.1. These results validate the importance of each part of the FPNN in enhancing model prediction accuracy, stability, and robustness, thereby supporting the rationality of its architectural design.

\begin{table}[H]
  \centering
  \caption{Ablation Experiments of FPNN at Optimal NOI}
  \begin{tabularx}{\textwidth}{XXXXX}
    \toprule
    Dataset & Detach & MAPE (\%) & MAE (cycles) & RMSE (cycles) \\
    \midrule
    10 Cycles \newline  & Initial layers & 3.45 & 25.43 & 37.73 \\
     & 3D conv & 3.09 & 23.27 & 34.67 \\
     & Residual & 2.58 & 19.32 & 26.67 \\
     & A branch & 98.92 & 772.33 & 834.04 \\
     & No detach & 2.47 & 18.09 & 26.49 \\
    \midrule
    20 Cycles \newline  & Initial layers & NaN & NaN & NaN \\
     & 3D conv & 1.66 & 12.82 & 20.74 \\
     & Residual & 2.32 & 17.52 & 26.98 \\
     & A branch & 96.55 & 771.39 & 844.44 \\
     & No detach & 1.29 & 10.60 & 17.60 \\
    \midrule
    30 Cycles \newline  & Initial layers & NaN & NaN & NaN \\
     & 3D conv & 1.45 & 11.07 & 18.93 \\
     & Residual & 1.23 & 9.20 & 13.11 \\
     & A branch & 93.72 & 749.21 & 817.94 \\
     & No detach & 1.08 & 8.04 & 12.48 \\
    \midrule
    40 Cycles \newline  & Initial layers & NaN & NaN & NaN \\
     & 3D conv & 1.04 & 8.05 & 12.16 \\
     & Residual & 0.81 & 6.31 & 9.47 \\
     & A branch & 90.63 & 714.83 & 775.20 \\
     & No detach & 0.88 & 6.81 & 10.03 \\
    \bottomrule
  \end{tabularx}
  \smallskip 
  \textbf{Note:} (1) 'NaN' indicates that during the training process of FPNN, after initial iterations, the loss value exhibits NaN anomalies. This may be due to the max pooling layer in the initial layers significantly reducing the feature dimensions, thereby making the model training simpler and more stable. (2) 'A branch' refers to the differential feature branch that is removed in the dual-stream network.
\end{table}

\begin{figure}[H]
  \centering
  \includegraphicsfit{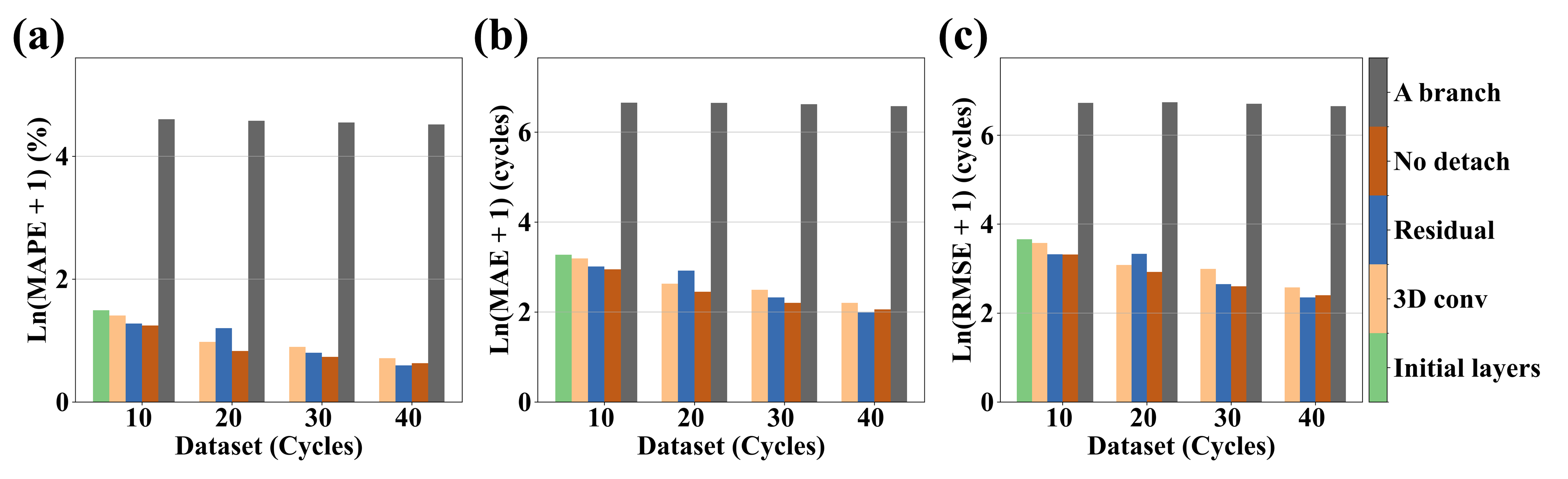}
  \caption{Results of the FPNN Ablation Experiments: (a) MAPE; (b) MAE; (c) RMSE.}
  \label{fig:image5}
\end{figure}

\subsection{ Model Interpretability Analysis}
To examine the interpretability of the model, this paper analyzes and demonstrates the internal mechanism of the FPNN model in handling EPBL tasks.

In FPNN, the 3D CNN layer is responsible for preprocessing features, providing suitable inputs for subsequent InceptionBlocks. The InceptionBlock structure contains four branches, each composed of a specific sequence of convolutional layers, undertaking different functions. For example, in the branch1×1, the 1×1 convolutional layer primarily serves the purpose of channel number conversion and integration of feature information, while also reducing the model's parameter count and enhancing computational efficiency. Within the InceptionBlock, most branches start with a 1×1 convolution kernel, aimed at rapidly adjusting the channel number of the feature map, preparing for deeper feature extraction. This swift channel transition facilitates the fusion of information across different channels, strengthening the model's adaptability and generalization ability to input data. The neural network learns features from data by optimizing weight parameters, in order to fit the target function.

Through weight visualization, one can observe the distribution of weights learned by the convolutional layers in each branch, providing insights into the model's decision-making process. Figure 6 displays the visualization of the convolutional layer weights in the InceptionBlock branches of the FPNN trained with the first10 cycles of data. As shown in Figure 6a-e, the weights of the 1×1 convolutional kernels vary in size, revealing how the FPNN differentially processes information across various channels. The 1×1 convolutional layer following the pooling layer in the branch\_pool highlights its ability to effectively transform features while reducing dimensions.

Figure 6f shows the distribution of the 3×3 convolution kernel weights in the branch3×3, which, compared to the 1×1 kernels, exhibit a more specialized feature extraction capability. The FPNN reduces reliance on a single set of features in this way, enhancing the model's generalization ability. In Figure 6g-h, the convolutional layers in the branch3×3stack branch achieve a receptive field equivalent to that of larger convolutional kernels by stacking multiple 3×3 layers, while maintaining a lower number of parameters.

Ultimately, through multi-scale feature extraction, the FPNN is capable of integrating complex nonlinear features extracted from different branches and linear features transmitted via residual connections, achieving effective feature fusion. The combination of multiple InceptionBlocks forming InceptionBlocks further extracts a broader range of advanced electrochemical features, significantly enhancing the predictive performance of the FPNN. This section provides a demonstrative analysis of the learning outcomes of all layers in the flexible unit InceptionBlock, indicating that the FPNN model is interpretable.

\begin{figure}[H]
  \centering
  \includegraphicsfit{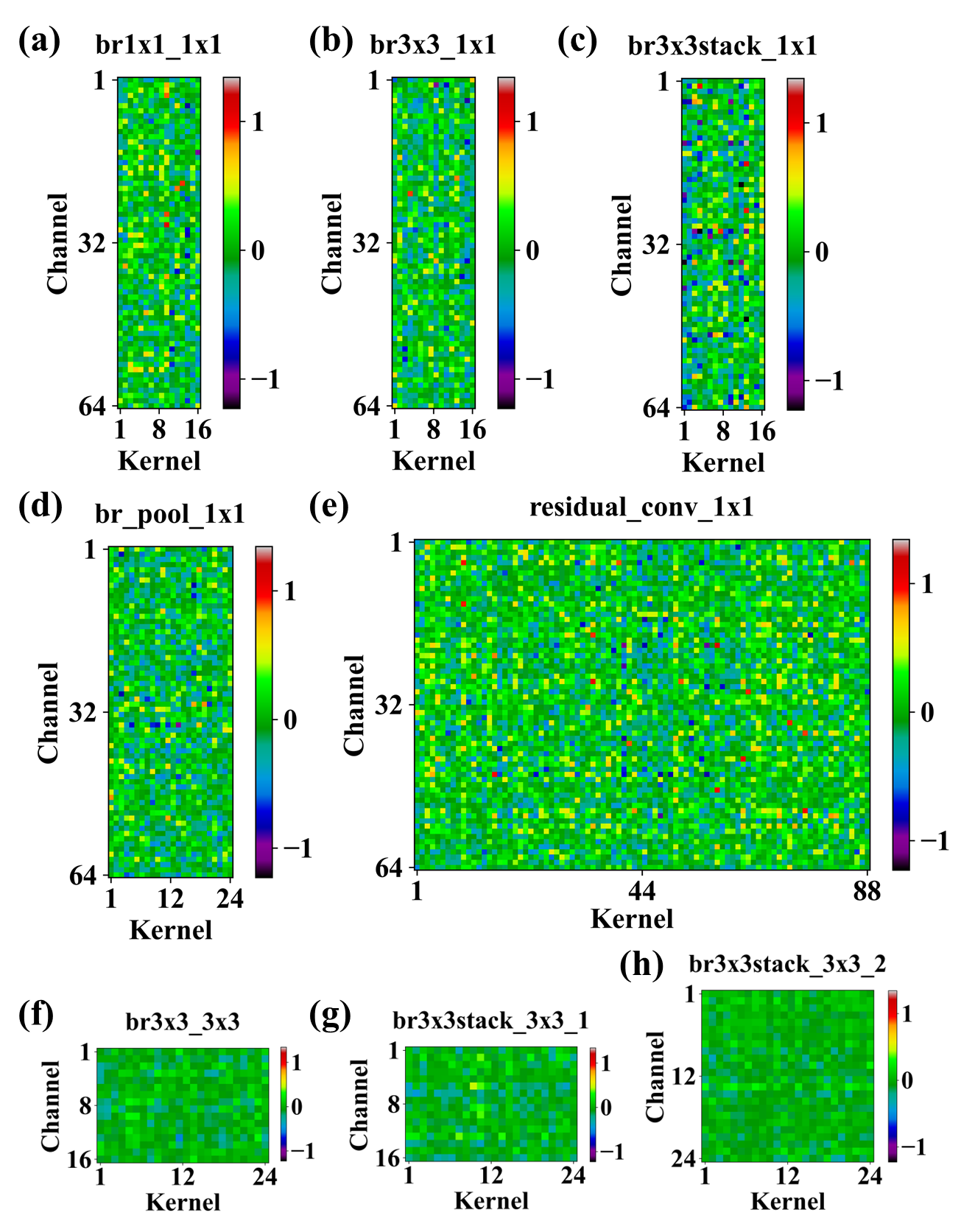}
  \caption{shows the heatmaps of the convolutional layer weights in the first InceptionBlock of the FPNN model trained with the first 10 cycles of data, revealing the role of different convolutional layers in feature extraction: (a) the convolutional layer with 1×1 kernels in branch1×1; (b) the convolutional layer with 1×1 kernels in branch3×3; (c) the convolutional layer with 1×1 kernels in branch3×3stack; (d) the convolutional layer with 1×1 kernels in branch\_pool; (e) the convolutional layer with 1×1 kernels in residual\_conv; (f) the convolutional layer with 3×3 kernels in branch3×3; (g) the first layer of the convolutional layer with 3×3 kernels in branch3×3stack; (h) the second layer of the convolutional layer with 3×3 kernels in branch3×3stack.}
  \label{fig:image6}
\end{figure}

\section{ Conclusion}
The FPNN model proposed in this paper, by integrating multiple modules such as InceptionBlock, 3D CNN, 2D CNN, and dual-stream networks, enhances the capability to extract electrochemical features of batteries from video-like format data. The FPNN model demonstrates high adaptability to different EPBL tasks by adaptively adjusting the number of InceptionBlocks. On the MIT dataset, the FPNN model achieves outstanding predictive accuracy in EPBL tasks, with MAPEs of 2.47\%, 1.29\%, 1.08\%, and 0.88\% when the input cyclic data volumes are 10, 20, 30, and 40, respectively. The study finds that with smaller data volumes, the FPNN tends to utilize more complex structures for feature extraction, reflected in higher optimal NOI values. As the data volume increases, the optimal NOI value gradually decreases, indicating that a more concise FPNN structure is sufficient for the task, and additional complexity may lead to a decline in predictive performance. Furthermore, ablation experiments further confirm the importance and necessity of each component in the FPNN architecture. Finally, through visual analysis of the weights of the internal layers of the InceptionBlock, this study enhances the interpretability of the model and highlights its effectiveness in complex feature extraction. Considering the commonalities between EPBL tasks and other machine learning tasks in the battery field, these characteristics of the FPNN suggest its potential advantages in broader applications within the battery domain.
\bibliography{references}

\end{document}